\documentclass[letterpaper, 10 pt, conference]{ieeeconf}  

\IEEEoverridecommandlockouts                              

\usepackage{graphics} 								
\usepackage{epsfig} 									
\usepackage{mathptmx} 								
\usepackage{times} 									
\usepackage{amsmath} 								
\usepackage{amssymb,amsfonts} 					
\usepackage{flushend} 								
\usepackage{cite}
\usepackage{placeins}
\usepackage{graphicx}
\usepackage{color,soul}
\usepackage{graphicx}
\usepackage{algorithm2e}
\usepackage{color}
\usepackage{graphics}
\usepackage{verbatim}
\usepackage{array}
\usepackage{wrapfig}
\usepackage{flushend}
\usepackage{subfigure}
\usepackage{paralist}
\usepackage{mathtools}

\usepackage{todonotes}

\title{\LARGE \bf
Comparing merging behaviors observed in naturalistic data with behaviors generated by a machine learned model
}

\author{Aravinda Ramakrishnan Srinivasan$^{1*}$, Mohamed Hasan$^{2}$, Yi-Shin Lin$^{1}$, Matteo Leonetti$^{2}$, Jac Billington$^{3}$, \\ Richard Romano$^{1}$, and Gustav Markkula$^{1*}$
\thanks{This project has received funding from UK Engineering and Physical Sciences Research Council under fellowship named COMMOTIONS - Computational Models of Traffic Interactions for Testing of Automated Vehicles - EP/S005056/1}
\thanks{$^{1}$ Institute for Transport Studies, University of Leeds, UK}%
\thanks{$^{2}$ School of Computing, University of Leeds, UK}%
\thanks{$^{3}$ School of Psychology, University of Leeds, UK}%
\thanks{$^{*}$ Corresponding authors: {\tt\small A.R.Srinivasan@leeds.ac.uk, G.Markkula@leeds.ac.uk}}%
}

\begin{document}
\maketitle
\thispagestyle{empty}
\pagestyle{empty}

\begin{abstract}
There is quickly growing literature on machine-learned models that predict human driving trajectories in road traffic. These models focus their learning on low-dimensional error metrics, for example average distance between model-generated and observed trajectories. Such metrics permit relative comparison of models, but do not provide clearly interpretable information on how close to human behavior the models actually come, for example in terms of higher-level behavior phenomena that are known to be present in human driving. We study highway driving as an example scenario, and introduce metrics to quantitatively demonstrate the presence, in a naturalistic dataset, of two familiar behavioral phenomena: (1) The kinematics-dependent contest, between on-highway and on-ramp vehicles, of who passes the merging point first. (2) Courtesy lane changes away from the outermost lane, to leave space for a merging vehicle. Applying the exact same metrics to the output of a state-of-the-art machine-learned model, we show that the model is capable of reproducing the former phenomenon, but not the latter. We argue that this type of behavioral analysis provides information that is not available from conventional model-fitting metrics, and that it may be useful to analyze (and possibly fit) models also based on these types of behavioral criteria.
\end{abstract}

\section{INTRODUCTION}
There is an increasing presence of vehicles with autonomous capabilities on the roadways \cite{ono2016pre,dikmen2016autonomous}. 
As road users share the road space, \emph{interactions} occur, in the form of situations \lq\lq where the behavior of at least two road users can be interpreted as being influenced by a space-sharing conflict between the road users\rq\rq~\cite{markkula2020interaction}. For these interactions between autonomous vehicles and other road users to be natural and safe, the autonomous vehicles need to understand other road users and anticipate their behaviors, and for this reason road user behaviour has been modeled at various levels, ranging from making a prediction about when a pedestrian will cross \cite{camara2020pedestrian,camara2020pedestrian2} to deriving driving models for different human driven vehicles \cite{claussmann2019review}. 

From a machine learned modeling perspective, predicting other road user behaviors can be formulated as the problem of predicting their trajectories. In recent years, recurrent neural networks (RNNs), a class of deep neural networks, have been used for these prediction tasks due to their ability to model time series data, and not least
Long Short-Term Memory (LSTM) based RNN models have been used to predict human trajectories while taking into account the neighbor's trajectories \cite{Alahi_2016_CVPR}. Deo and Trivedi expanded on this idea for vehicle trajectory prediction, by utilizing a convolutional layer to preserve the spatial relationship between neighboring vehicles and predicted trajectory for the vehicle of interest \cite{deo2018convolutional}. This convolutional social pooling (CSP) LSTM algorithm has since been used as a benchmark to measure the accuracy of newer models \cite{Ma_Zhu_Zhang_Yang_Wang_Manocha_2019,8917228}.  Mozaffari et al. \cite{mozaffari2020deep} present a comprehensive literature review of deep-learning-based prediction algorithms. 

The most common metric used in the machine learning literature to support the performance of a trajectory prediction algorithm is the root mean square error (RMSE) between the prediction and the actual trajectory; either the average displacement error over the entire prediction horizon or the the final displacement error. In models that give probability over the different maneuvers, negative log likelihood (NLL) is also reported. The drawback in simplifying the performance of the model to a single quantitative value is loss of higher-level, qualitative context about the different types of behaviors actually exhibited by the models. Conventional comparisons between different machine learned models can indicate which model reproduces the human trajectories more closely, but how low RMSE or NLL values are low enough? Do the machine learned models actually reproduce the higher level behaviors exhibited by humans?

Highway merging of vehicles can be considered as a microcosm of the complex interactions that happen during everyday driving and might thus serve as a valid example scenario to test the ability of the machine learned model to successfully navigate in traffic. 
During highway merging, space-sharing conflicts are common between the on-ramp vehicles and vehicles on the outermost lane of the highway, and the interaction between human drivers in this type of situation has been the topic of substantial white box (non machine-learned) modeling in the past \cite{choudhury_modeling_2009,kang2017game}. Implicitly, these models suggest that when one vehicle has a kinematic lead over the other, it will tend to pass the merging point first, but in situations where it is less clear that one agent is kinematically leading, who will go first is harder to predict, and will in practice depend on a competition (or cooperation) between the drivers.
Another behavior that has been studied and modeled extensively in literature is the lane change behavior of the highway vehicle and the merging vehicle \cite{yu2018human,zheng2014recent}. The lane changes were attributed to different goals like, highway vehicles preferring to avoid the deceleration to accommodate the merging vehicles \cite{kondyli2009driver}, the merging vehicle forcing the lane change due to end of on-ramp or driver's aggressive preference, or the merging vehicle finding suitable gap in the traffic to merge safely \cite{choudhury_modeling_2009}. Here we specifically study the "courtesy" lane change behavior of highway vehicles to accommodate the on-ramp vehicles.   
As far as we are aware, neither of these two phenomena (tendency of kinematically leading agents to pass first in merging situations, and tendency of on-highway vehicles to change lane to provide space for on-ramp vehicles) have been explicitly studied in naturalistic data, let alone in the behavior of machine-learned driver models.  

Thus, the primary aim of this work is to outline an approach for obtaining richer insights--compared to RMSE or NLL--about the output of machine-learned models.
Specifically, we propose analysis methods designed for identifying the two above-mentioned behavioral phenomena in naturalistic data, and then apply the same methods to the behavior predictions of one of the benchmark RNN models, the CSP-LSTM of Deo and Trivedi \cite{deo2018convolutional}. This model was chosen due to the availability of the code\cite{CSP-LSTMcode} and the dataset \cite{NGSiMData} used. 

\section{METHODS}
 
\subsection{Machine-learned model}
\begin{figure}[]
      \centering
      \includegraphics[scale=0.27]{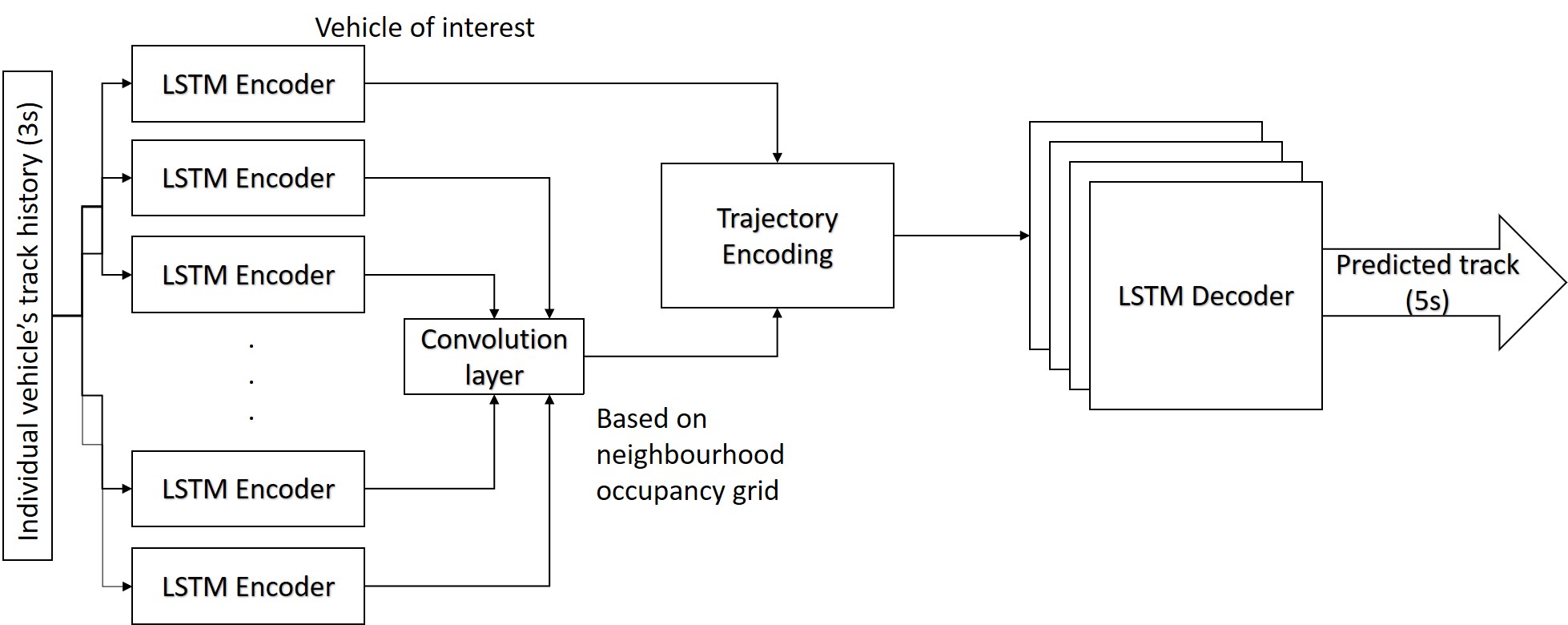}
      \caption{CSP-LSTM network architecture block diagram.}
      \label{fig:CSP-LSTM}
\end{figure}

Deo and Trivedi \cite{deo2018convolutional} introduced a CSP-LSTM architecture for vehicle trajectory prediction, as shown in Fig. \ref{fig:CSP-LSTM}. The model takes as input $t_{CSP_{ip}} = 3$ seconds of trajectory data for both the vehicle of interest and the surrounding vehicles. It encodes each vehicle's trajectory with a LSTM based encoder. A convolutional layer is utilized to preserve the spatial relationship between the surrounding vehicles' encoded trajectories. The encoded trajectory of the vehicle of interest along with the convolutional layer output for the surrounding vehicles are passed through a LSTM-based decoder layer which produces a $5$ seconds trajectory prediction for the vehicle of interest. This can then be repeated for each vehicle in a given driving scene.

\subsection{Dataset}

Deo and Trivedi \cite{deo2018convolutional} utilized the NGSIM dataset \cite{NGSiMData} for training their CSP-LSTM network. The NGSIM dataset consists of human driven vehicle trajectories from two different highways in USA, the US101 and I80 highways. 
We reproduced the same model fitting regime as Deo and Trivedi, with exactly the same 70-10-20 split of the dataset into training, validation, and test sets, respectively, and we verified that our trained network achieved the same RMSE performance as reported in \cite{deo2018convolutional}. In this paper, all the presented and depicted behavioral comparisons were made between the naturalistic trajectories and the machine learned models output for only the validation and the test splits of the dataset. There was a total of 1,667 unique vehicles on the US101 and 1,268 unique vehicles on the I80 in the test and validation set. Out of these, $111$ and $147$ vehicles merged onto the highways from the on-ramp respectively. 

\subsection{Behavioral analysis}

Humans are known to rely a lot on the first-order motion information when judging collision conflicts \cite{markkula2016farewell}. It is also known that humans have a fundamental ability to estimate the time-to-arrival (TTA) of approaching objects \cite{lee1976theory}. For these reasons we hypothesized that we would be able to observe salient behavioral patterns by analysing only the first-order kinematics. Let, $t_m$ be the time the merge actually happened and $\tau$ be the look back window. From the naturalistic trajectories, the time, $t_m$ and position of each merging scenario can be extracted. Since, we are interested in understanding what contributed to the merge happening the way it had happened, the look back window is essentially the length of history before the merge happened, which we utilize to help our behavioral understanding/analysis. Since, CSP-LSTM was capable of predicting up to $5$ seconds of trajectory, the history or look back window, $\tau$ was chosen from $1$ second up to $5$ seconds at $1$ second intervals before the actual merge ($t_m$) happened.

For both the vehicle on the highway and the on-ramp vehicle, the interacting pair, it is straightforward to find their distance to merging point at time, $t_m - \tau$,  and also compute the instantaneous velocities of the vehicle from the naturalistic driving data. Let, $v^h_t, d^h_t$ and  $v^m_t, d^m_t$ denote the instantaneous velocity and distance to the merging point for the highway vehicle and the on-ramp merging vehicle respectively at a given time, $t$. We can then compute the lead time for the highway vehicle for a given look back window $\tau$, $T_\tau$ with the formula $T_\tau = d^h_{t_m - \tau}/v^h_{t_m - \tau} - d^m_{t_m - \tau}/v^m_{t_m - \tau}$.

\begin{figure}[]
      \centering
      \includegraphics[scale=0.27]{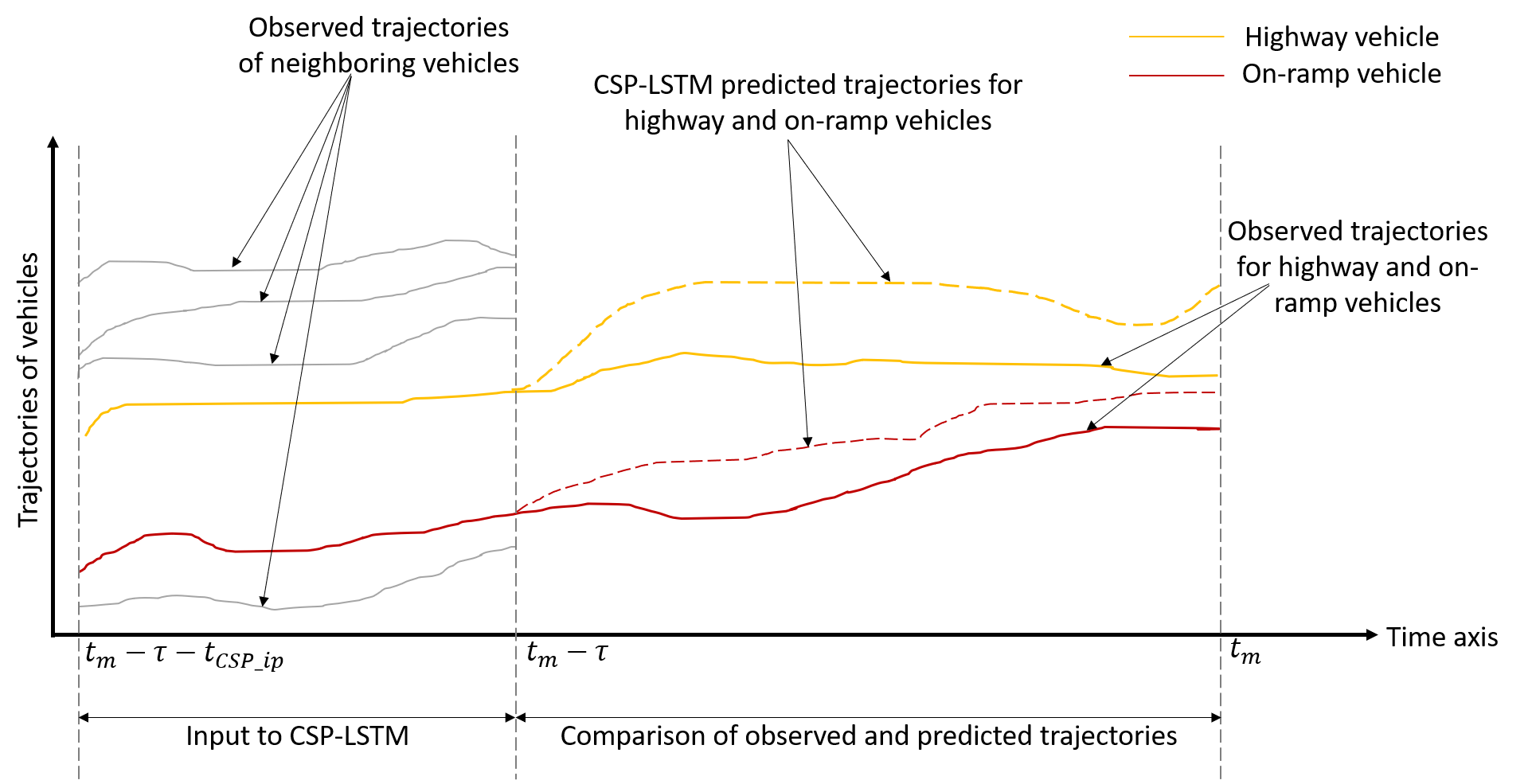}
      \caption{Schematics of behavior comparison between naturalistic and CSP-LSTM generated trajectories.}
      \label{fig:CSP-LSTM-sche}
\end{figure}

The behaviors of interest were analyzed for the naturalistic data and the machine-learned model with the exact same procedure utilizing the lead time for the highway vehicle. A schematic of the behavioral comparison is shown in fig. \ref{fig:CSP-LSTM-sche}. The trajectories has been represented in one (lower) dimension. In reality, both the lateral and longitudinal positions of the vehicles were utilized for training the machine-learned model and the behavioral analysis. The trained CSP-LSTM model was given trajectory history of $3$ seconds as inputs from appropriate time stamp for the vehicle of interest (the highway vehicle in the interacting pair) and neighboring vehicles. The output of the trained model, the $5$ seconds predicted trajectory of the vehicle of interest was analyzed for the behavior exhibited (fig. \ref{fig:CSP-LSTM-sche}). 

\subsubsection{Bias for kinematically leading agent to pass first}

Fig. \ref{fig:KLA_illu} shows a schematic example for the bias for kinematically leading agent to pass first behavior.
Given a lead time for the highway vehicle, a positive value indicates that highway vehicle had an apparent lead over the merging on-ramp vehicle kinematically at that particular look back time and negative value indicates that the merging on-ramp vehicle had a lead. We would expect that when the absolute apparent lead $|T_{\tau}|$ is large enough, the kinematically leading agent would always pass the merging point first. Additionally, we would expect this pattern to break down as $|T_{\tau}|$ approaches zero. 


\subsubsection{Changing lane to yield}
\begin{figure}[h]
    
      \centering
      \subfigure[]{
      \includegraphics[scale=0.25]{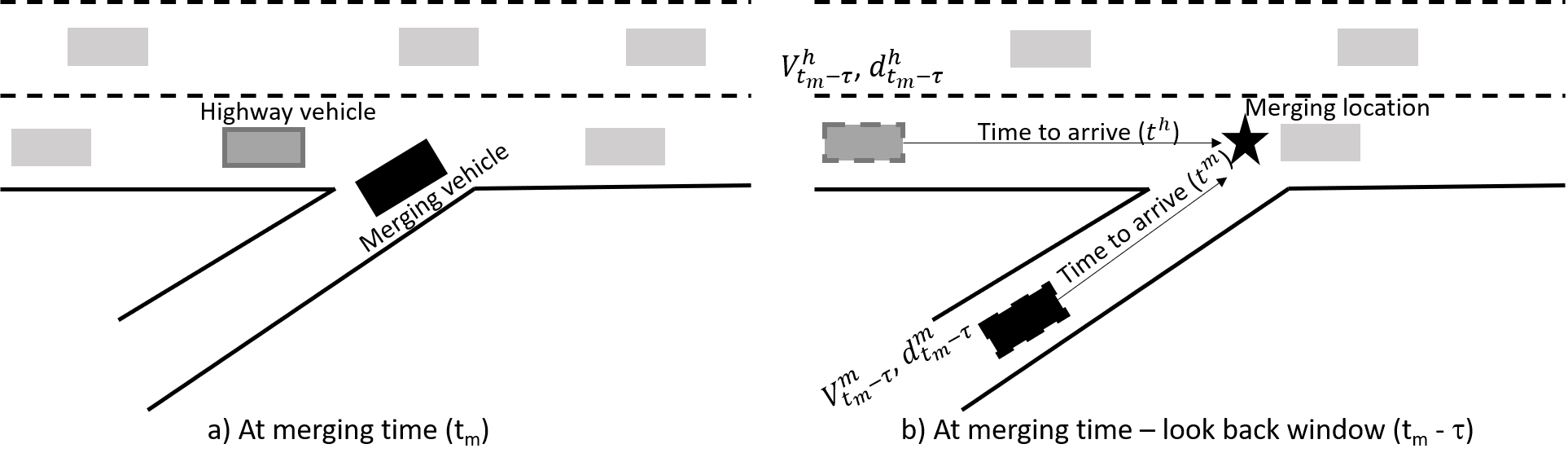}\label{fig:KLA_illu}}
      \subfigure[]{
      \includegraphics[scale=0.25]{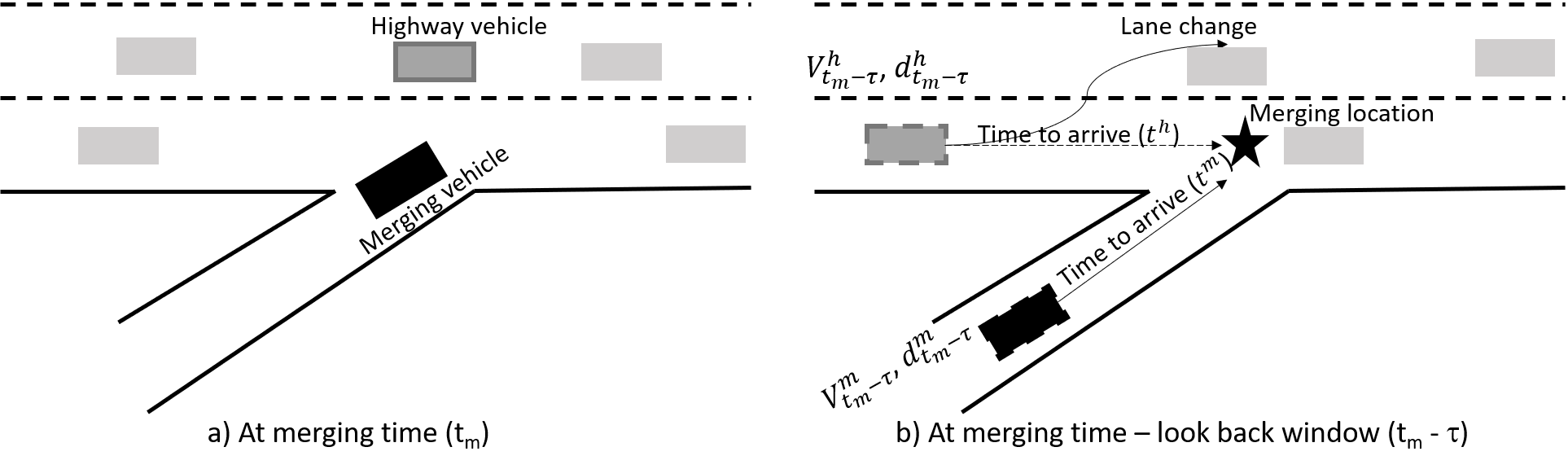}\label{fig:LC_illu}}
      \caption{A pictorial representation of behavioral phenomeana a) Bias for kinematically leading agent to pass the merging point first b) Lane changing by highway vehicle to accommodate the on-ramp vehicle}
      
\end{figure}

Fig. \ref{fig:LC_illu} illustrates a hypothetical lane change that happened between the merging time, $t_m$ and the look back window, $\tau$ to accommodate the on-ramp vehicle merging into the highway. For the lane change behavior statistics, we count all lane changes which happened between the look-back window $t_m-\tau$ and the merging time $t_m$, since these can be considered as a potential courtesy lane changes. We expect the frequency of lane changes to increase when there is imminent space-sharing conflict, i.e., for small $|T_{\tau}|$. 

\section{RESULTS}

\begin{figure*}[]
      \centering
      \subfigure[]{
      \includegraphics[scale = 0.45]{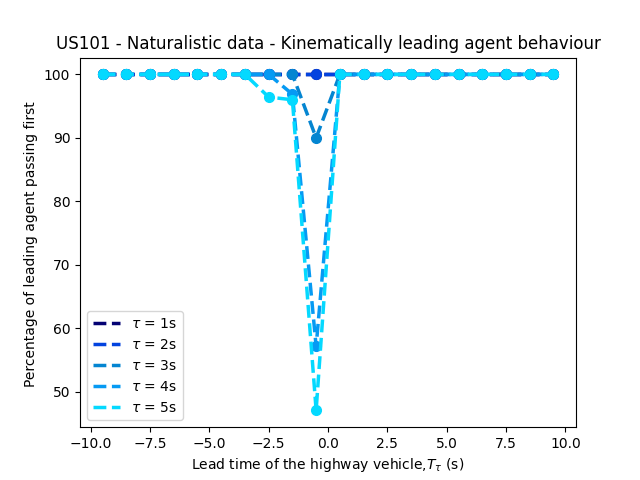} \label{fig:KLA_us101_nat}}
      \subfigure[]{
      \includegraphics[scale=0.45]{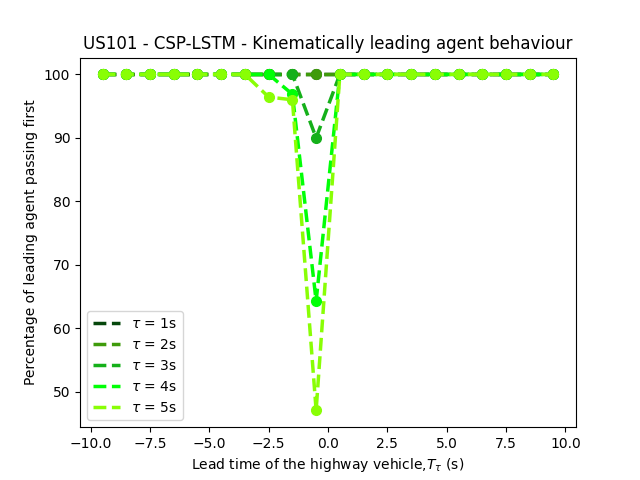}\label{fig:KLA_us101_csp}}
      \subfigure[]{
      \includegraphics[scale=0.45]{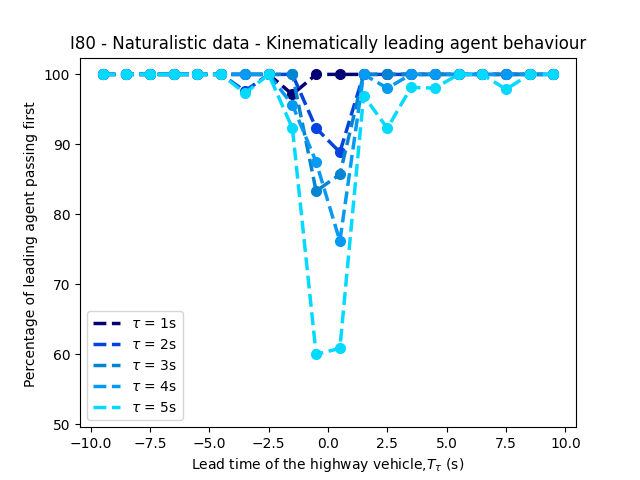}\label{fig:KLA_i80_nat}}
      \subfigure[]{
      \includegraphics[scale=0.45]{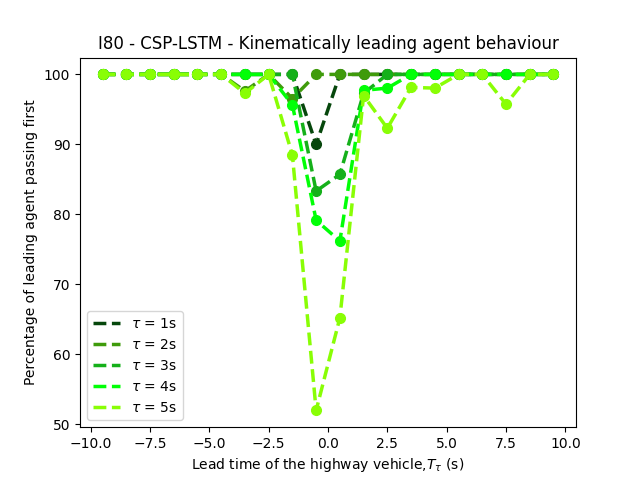}\label{fig:KLA_i80_csp}}
      
      \caption{Bias for kinematically leading agent to pass first behavior statistics}
      \label{fig:KLA}
   \end{figure*}
\subsection{Bias for kinematically leading agent to pass first}

 The frequencies of the initially kinematically leading agent passing the merging point first, as exhibited by the human driven vehicles for the US101 and I80 highways are shown in fig. \ref{fig:KLA_us101_nat} and fig. \ref{fig:KLA_i80_nat} respectively, as functions of the look-back time $\tau$ and the apparent kinematic lead $T_{\tau}$ for the highway vehicle at the look-back time. There are two obvious patterns: First, in line with our expectations, when either vehicle had a clear kinematic lead over the other, they ended up passing the merging point first; as soon as $|T_{\tau}|$ is above about 1 s for the (higher speed) US101, or above 2 s for the (lower speed) I80, the frequency of the leading agent passing first is at or close to 100\%. Second, when there is imminent space-sharing conflict, that is when $|T_{\tau}|$ is closer to $0$ seconds, there is a gradual shift to lower percentages, indicating a further level of interaction. This is especially so for larger $\tau$, indicating--quite naturally--that the ultimate outcome of a close merging cooperation/competition may be harder to predict from snapshots further into the past.  
 
 Comparing the behaviors exhibited by the naturalistic data, fig. \ref{fig:KLA_us101_nat} and \ref{fig:KLA_i80_nat} with the CSP-LSTM model generated behaviors, fig. \ref{fig:KLA_us101_csp} and \ref{fig:KLA_i80_csp}, it is clear that the CSP-LSTM model was able to produce trajectories which exhibited very similar levels of bias for the kinematically leading agent to pass the merging point first. Thus, at this level of analysis, the RMSE focused learning algorithm was enough to learn and reproduce this behavior pattern.

\subsection{Changing lane to yield}
\begin{figure*}[]
      \centering
      \includegraphics[scale=0.35]{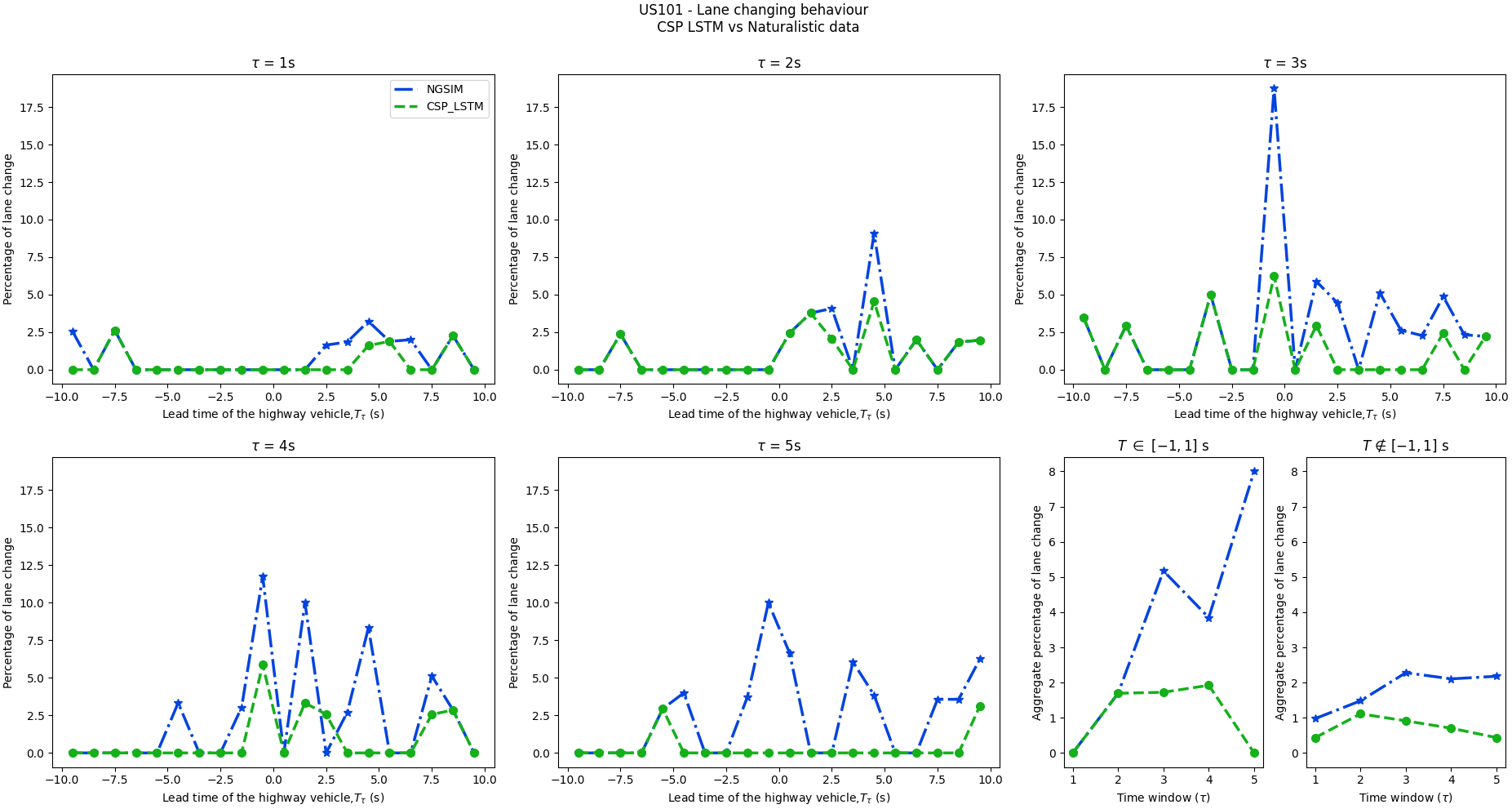}
      \caption{US101 - Lane change analysis for the naturalistic driving and trajectories generated by the CSP-LSTM.}
      \label{fig:LC_us101}
\end{figure*}
\begin{figure*}[]
      \centering
      \includegraphics[scale=0.35]{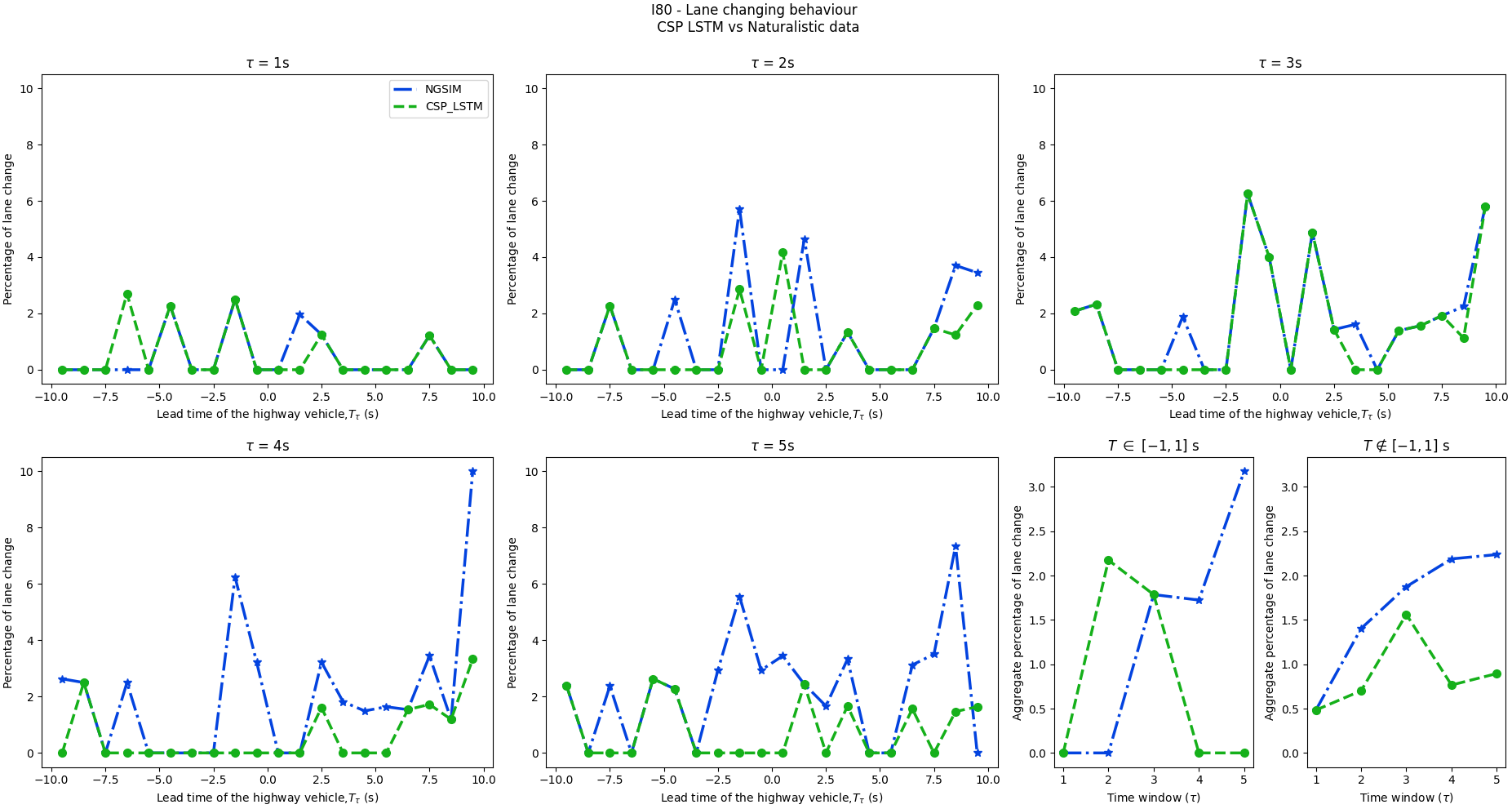}
      \caption{I80 - Lane change analysis for the naturalistic driving and trajectories generated by the CSP-LSTM.}
      \label{fig:LC_i80}
\end{figure*}
\begin{figure*}[]
      \centering
      \subfigure[]{
      \includegraphics[scale=0.45]{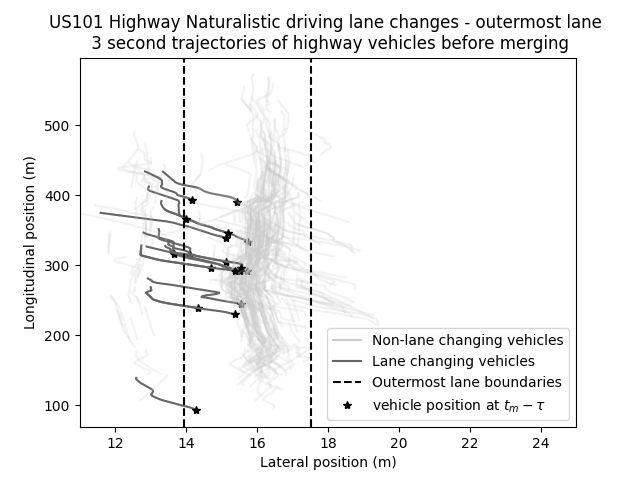}}
      \subfigure[]{
      \includegraphics[scale=0.45]{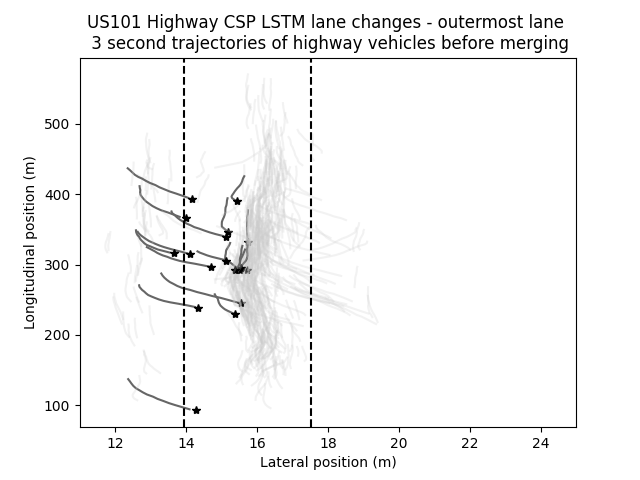}}
      \subfigure[]{
      \includegraphics[scale=0.45]{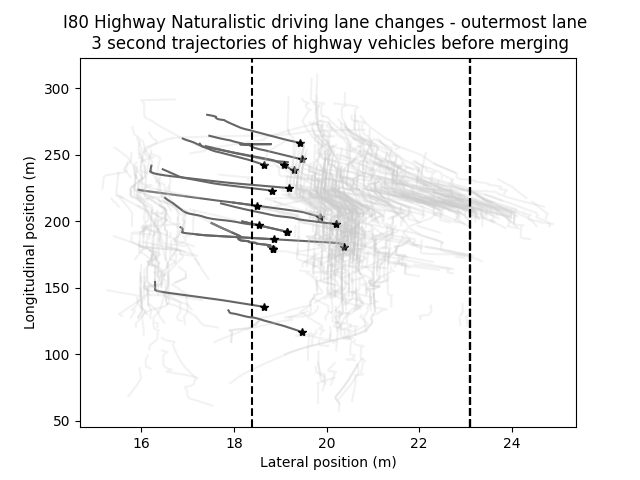}}
      \subfigure[]{
      \includegraphics[scale=0.45]{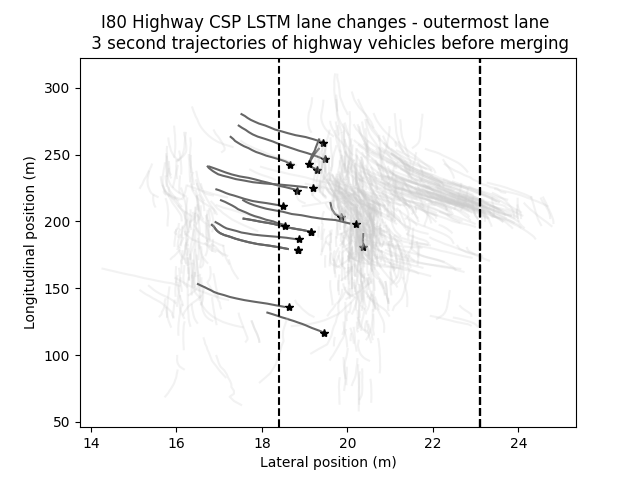}}
      
      \caption{Naturalistic and CSP-LSTM generated vehicle trajectories in potential lane changing scenarios for the vehicle on the outermost lane of the highway to accommodate the merging vehicle.}
      \label{fig:LC_traj}
   \end{figure*}
Fig. \ref{fig:LC_us101} and \ref{fig:LC_i80} show the prevalence of lane change behavior in the naturalistic driving and the CSP-LSTM generated trajectories in US101 and I80 respectively. These observations are somewhat noisy due to the limited sample size, but is nevertheless clear that in the US101 highway data (fig. \ref{fig:LC_us101}), when there is a space sharing conflict and there is sufficient time to successfully change lanes, $\tau \geq 3$ s there is a spike in lane change occurrences. However, the CSP-LSTM generated trajectories does not reproduce this increase in frequency of lane changes. The same data is shown in summary form in the bottom right panes of the fig. \ref{fig:LC_us101}, indicating that human drivers in space-sharing conflict situations (lead time for highway vehicle $T_\tau$ $\in$ $[-1,1]$ s), show an increased tendency to switch lanes, presumably out of courtesy to avoid the conflict, whereas the CSP-LSTM predicted trajectories do not reproduce this pattern.  Also in non-space sharing conflict situations ($|T_\tau| > 1 s$), human and CSP-LSTM lane changing frequencies are not matching.  
With the I80 highway (fig. \ref{fig:LC_i80}), the lane change peaks are not as clear as with the US101 highway in the naturalistic data (fig. \ref{fig:LC_i80}). Nevertheless, in the space-sharing conflict zone, the bottom right corner in fig. \ref{fig:LC_us101} and fig. \ref{fig:LC_i80}, the lane changing behaviors are similar to one another. The analyzes when done at the space-sharing conflict zone and non-space conflict zone, clearly show that the CSP-LSTM is not faithfully reproducing the lane changing behavior exhibited by the humans. 
\section{DISCUSSION AND CONCLUSION}

This paper presented a new, arguably more human-centric, way to analyze the capabilities of machine-learned models which are used in the domain of autonomous driving for predicting vehicle trajectories. To illustrate the proposed approach, two example behavioral phenomena in highway merging were targeted, analysis methods were developed to demonstrate their presence in naturalistic data, and subsequently also applied to the model's predictions. The results showed that the machine-learned model was able to reproduce well a phenomenon whereby the kinematically leading agent passes the merging point first when the apparent kinematic lead is large ($>1-2$ s), up to $5$ seconds in advance of the merge, and where this predictability also deteriorates for less clear kinematic leads. Thus, the simplification of the road user behavior learning to a pure trajectory learning task seems to be justified for replicating the kinematically leading agent bias to pass the merging point first as observed in the naturalistic driving data. 

The other targeted phenomenon was courtesy lane changing behavior by the highway vehicles, to facilitate the merging of on-ramp vehicles. In short, the machine learned model was not able to reproduce this phenomenon well. To further understand the human and model lane changing behavior, the trajectories of the vehicles in both highways from the naturalistic and CSP-LSTM generated data are shown in fig. \ref{fig:LC_traj}. It is clear from this figure that the limitation of the CSP-LSTM model does not lie in a general inability of predicting lane changing behavior; it clearly does predict some of the observed naturalistic lane changes. However, it does in general underpredict their frequency, and, crucially, it is clear from the bottom right panes of especially fig. \ref{fig:LC_us101} and \ref{fig:LC_i80}, the model does not capture the context-sensitive increased tendency of human drivers to change lanes when there is an apparent space-sharing conflict with a merging vehicle.

Overall, the findings presented here demonstrate how a richer analysis of human and model-predicted behavior can provide a better understanding of the capabilities of machine-learned models. It is clear that the CSP-LSTM model is capable of capturing some advanced behavioral phenomena in impressive detail, yet is unable to capture other phenomena. Notably, neither of these insights are accessible from conventional performance metrics such as RMSE or NLL. Our results also open for many interesting future research opportunities: Our overall analysis approach can be generalized to a wider range of salient behavioral phenomena, across a wider range of interaction scenarios \cite{markkula2020interaction}. The added insights into the behavioral capabilities of the machine-learned models may be leveraged both in algorithms making use of the models, and to develop improved learning targets for the models, to help ensure that future models can be more behaviorally competent.


\bibliographystyle{IEEEtran}
\bibliography{Srinivasan_Comparing_merging_behavior_2021.bib}

\end{document}